\pdfoutput=1

\documentclass[11pt]{article}

\usepackage{acl}

\usepackage{times}
\usepackage{latexsym}

\usepackage[T1]{fontenc}

\usepackage[utf8]{inputenc}
\usepackage{enumitem}
\usepackage{microtype}
\usepackage{amsmath}
\usepackage{graphicx}
\usepackage{multirow}
\usepackage{soul}

\usepackage{times}
\usepackage{latexsym}
\usepackage{graphicx}
\usepackage{booktabs}
\usepackage{subcaption}
\usepackage[ruled,vlined]{algorithm2e}
\usepackage{multirow}
\usepackage{tabularx}
\usepackage{amsmath}
\usepackage{bbold}
\usepackage{mathtools}
\usepackage{xcolor}
\usepackage{tabularx,ragged2e}
\usepackage{microtype}

\title{Do Selective Prediction Approaches Consistently Outperform the Simplest Baseline `\textit{MaxProb}'?}
\title{Selective Prediction Approaches Fail to Consistently Outperform the Simplest Baseline Despite Using Additional Resources}
\title{Investigating Selective Prediction Approaches Across Several Tasks in IID, OOD, and Adversarial Settings}

\author{Neeraj Varshney,~~ 
  Swaroop Mishra,~~ 
  Chitta Baral
  \\
  Arizona State University \\
  \texttt{\{nvarshn2, srmishr1, cbaral\}}@asu.edu
  }
  
\begin{document}
\maketitle
\begin{abstract}
In order to equip NLP systems with selective prediction capability, several task-specific approaches have been proposed.
However, which approaches work best across tasks or even if they consistently outperform the simplest baseline `MaxProb' remains to be explored.
To this end, we systematically study `selective prediction' in a large-scale setup of $17$ datasets across several NLP tasks.
Through comprehensive experiments under in-domain (IID), out-of-domain (OOD), and adversarial (ADV) settings, we show that despite leveraging additional resources (held-out data/computation), none of the existing approaches consistently and considerably outperforms MaxProb in all three settings.
Furthermore, their performance does not translate well across tasks.
For instance, \textit{Monte-Carlo Dropout} outperforms all other approaches on Duplicate Detection datasets but does not fare well on NLI datasets, especially in the OOD setting.
Thus, we recommend that future selective prediction approaches should be evaluated across tasks and settings for reliable estimation of their capabilities.

\end{abstract}

\section{Introduction}

Despite impressive progress made in Natural Language Processing (NLP), it is unreasonable to expect models to be perfect in their predictions.
They often make incorrect predictions, especially when inputs tend to diverge from their training data distribution \cite{elsahar-galle-2019-annotate,miller2020effect,pmlr-v139-koh21a}.
While this is acceptable for tolerant applications like movie recommendations, high risk associated with incorrect predictions hinders the adoption of these systems in real-world safety-critical domains like biomedical and autonomous robots.
In such scenarios, \textit{selective prediction} becomes crucial as it allows maintaining high accuracy by abstaining on instances where error is likely.

    

Selective Prediction (SP) has been studied in machine learning \cite{chow1957optimum,el2010foundations} and computer vision \cite{Geifman2017SelectiveCF, Geifman2019SelectiveNetAD}, but has only recently gained attention in NLP.
\citet{kamath-etal-2020-selective} proposed a post-hoc calibration-based SP technique for Question-Answering (QA) datasets. 
\citet{garg2021will} distill the QA model to filter out error-prone questions.
Unfortunately, despite the shared goal of making NLP systems robust and reliable for real-world applications, SP has remained underexplored; the community does not know which techniques work best across tasks/settings or even if they consistently outperform the simplest baseline `\textit{MaxProb}' \cite{hendrycks17baseline} (that uses the maximum softmax probability as the confidence estimator for selective prediction).


In this work, we address the above point and study selective prediction in a large-scale setup of $17$ datasets across NLI, Duplicate Detection, and QA tasks. 
We conduct comprehensive experiments under In-Domain (IID), Out-Of-Domain (OOD), and Adversarial (ADV) settings that result in the following findings:
\begin{enumerate}[noitemsep,nosep,leftmargin=*]
    \item None of the existing SP approaches consistently and considerably outperforms \textit{MaxProb}.
    
    \textbf{Slight improvement in IID:} Most of the approaches outperform MaxProb in the IID setting; however, the magnitude of improvement is very small (Figure \ref{fig:iid_results}). For instance, \textit{MCD} achieves an average improvement of just $0.28$ on AUC value across all NLI datasets.
    
    \textbf{Negligible improvement in OOD:} 
    The magnitude of improvement is even lesser ($0.08$) than that observed in the IID setting (Figure \ref{ood_fig}). 
    In a few cases, we also observe performance degradation (higher AUC than MaxProb). 
    
    \textbf{Performance degradation in ADV:} All the approaches fail to even match the MaxProb performance in ADV setting (Figure \ref{adv_fig}). For instance, \textit{MCD} degrades the AUC value by $1.76$ on duplicate detection datasets and \textit{calibration} degrades by $1.27$ on NLI datasets in ADV setting.

    \item \textbf{Approaches do not translate well across tasks}: We find that a single approach does not achieve the best performance across all tasks. For instance, \textit{MCD} outperforms all other approaches on Duplicate Detection datasets but does not fare well on the NLI datasets.  
    
    \item \textbf{Existing approaches require additional resources}: \textit{MCD} requires additional computation and \textit{calibration}-based approaches require a held-out dataset. 
    In contrast, \textit{MaxProb} does not require any such resources and still outperforms them, especially in the ADV setting.
    
\end{enumerate}

Overall, our results highlight that there is a need to develop stronger selective prediction approaches that perform well across tasks while being computationally efficient. 
To foster development in this field, we release our code and experimental setup.


\section{Selective Prediction}
\subsection{Formulation}
A selective prediction system comprises of a predictor ($f$) that gives the model's prediction on an input ($x$), and a selector ($g$) that determines if the system should output the prediction made by $f$ i.e.
\begin{equation*}
        (f,g)(x) =
            \begin{cases}
              f(x), & \text{if g(x) = 1} \\
              Abstain, & \text{if g(x) = 0}
            \end{cases}
\end{equation*}
Usually, $g$ comprises of a confidence estimator $\tilde{g}$ that indicates $f's$ prediction confidence and a threshold $th$ that controls the abstention level:
\begin{equation*}
    g(x) = \mathbb{1}[\tilde{g}(x)) > th]
\end{equation*}

An SP system makes trade-offs between $coverage$ and $risk$.
For a dataset $D$, coverage at a threshold $th$ is defined as the fraction of total instances answered by the system (where $\tilde{g} > th$) and risk is the error on the answered instances:
\begin{equation*}
    coverage_{th} = \frac{\sum_{x_i \in D} \mathbb{1}[\tilde{g}(x_i)) > th]}{|D|} 
\end{equation*}
\begin{equation*}
    risk_{th} = \frac{\sum_{x_i \in D} \mathbb{1}[\tilde{g}(x_i)) > th]l_i}{\sum_{x_i \in D} \mathbb{1}[\tilde{g}(x_i)) > th]} 
\end{equation*}
where, $l_i$ is the error on instance $x_i$.

With decrease in $th$, coverage will increase, but the risk will usually also increase.
The overall SP performance is measured by the \textit{area under Risk-Coverage curve} \cite{el2010foundations} which plots risk against coverage for all threshold values.
\textbf{Lower the AUC, the better the SP system} as it represents lower average risk across all thresholds.
We note that confidence calibration and OOD detection are related tasks but are non-trivially different from selective prediction as detailed in section \ref{sec:related_tasks}.

\subsection{Approaches}

Usually, the last layer of models has a softmax activation function that gives the probability distribution $P(y)$ over all possible answer candidates $Y$. $Y$ is the set of labels for classification tasks, answer options for multiple-choice QA, all input tokens (for start and end logits) for extractive QA, and all vocabulary tokens for generative tasks.
Thus, predictor $f$ is defined as: $\underset{y \in Y}{\mathrm{argmax}} P(y)$

\paragraph{Maximum Softmax Probability} (MaxProb):
\citet{hendrycks17baseline} introduced a simple method that uses the maximum softmax probability as the confidence estimator $\tilde{g}$ i.e. $\max_{y \in Y} P(y)$

\paragraph{Monte-Carlo Dropout} (MCD):
\citet{gal2016dropout} proposed to make multiple predictions on the test input using different dropout masks and ensemble them to get the confidence estimate.

\paragraph{Label Smoothing} (LS):
\citet{Szegedy2016RethinkingTI} proposed to compute cross-entropy loss with a weighted mixture of target labels during training instead of `hard' labels.
This prevents the network from becoming over-confident in its predictions.

\paragraph{Calibration} (Calib):
In calibration, a held-out dataset is annotated based on the correctness of the model's predictions (correct as positive and incorrect as negative) and another model (calibrator) is trained on this annotated binary classification dataset.
The softmax probability assigned to the positive class is used as the confidence estimator for SP.
\citet{kamath-etal-2020-selective} study a calibration-based SP technique for Question Answering datasets.
They train a random forest model as calibrator over features such as input length and probabilities of top 5 predictions.
We refer to this approach as \textbf{Calib C}.
Inspired by calibration technique presented in \citet{10.1162/tacl_a_00407}, we also train calibrator as a regression model (\textbf{Calib R}) by annotating the heldout dataset on a continuous scale instead of categorical labels (positive and negative as done in Calib C).
We compute these annotations using MaxProb as:
\begin{equation*}
        s =
            \begin{cases}
              0.5+ \frac{maxProb}{2}, & \text{if correct} \\
              0.5- \frac{maxProb}{2}, & \text{otherwise}
            \end{cases}
\end{equation*}
Furthermore, we train a transformer-based model for calibration (\textbf{Calib T}) that leverages the entire input text instead of features derived from it \cite{garg2021will}.


\section{Experimental Setup}
\label{sec_experimental_setup}
\subsection{Tasks and Settings:}
We conduct experiments with $17$ datasets across Natural Language Inference (NLI), Duplicate Detection, and Question-Answering (QA)s tasks and evaluate the efficacy of various SP techniques in IID, OOD, and adversarial (ADV) settings.

\textbf{NLI:} We train our models with SNLI \cite{bowman-etal-2015-large} / MNLI \cite{williams-etal-2018-broad} / DNLI \cite{welleck-etal-2019-dialogue} and use HANS \cite{mccoy-etal-2019-right}
, Breaking NLI \cite{glockner-etal-2018-breaking}, NLI-Diagnostics \cite{wang-etal-2018-glue}
, Stress Test \cite{naik2018stress}
as \underline{adversarial} datasets.
While training with SNLI, we consider SNLI evaluation dataset as \underline{IID} and MNLI, DNLI datasets as \underline{OOD}.
Similarly, while training with MNLI, we consider SNLI and DNLI datasets as OOD.

\textbf{Duplicate Detection:} We train with QQP \cite{iyer2017first} / MRPC \cite{dolan2005automatically} and use PAWS-QQP, PAWS-Wiki \cite{zhang-etal-2019-paws} as adversarial datasets.

\textbf{QA:} We train with SQuAD \cite{rajpurkar-etal-2016-squad} and evaluate on NewsQA \cite{trischler-etal-2017-newsqa}, TriviaQA \cite{joshi-etal-2017-triviaqa}, SearchQA \cite{dunn2017searchqa}, HotpotQA \cite{yang-etal-2018-hotpotqa}, and Natural
Questions \cite{kwiatkowski-etal-2019-natural}.

\subsection{Approaches:}
\quad{\textbf{Training: }}We run all our experiments using \textit{bert-base} model \cite{devlin-etal-2019-bert}
with batch size of $32$ and learning rate ranging in $\{1{-}5\}e{-}5$. 
All experiments are done with Nvidia V100 16GB GPUs.

\textbf{Calibration: }For calibrating QA models, we use input length, predicted answer length, and softmax probabilities of top $5$ predictions as the features (similar to \citet{kamath-etal-2020-selective}).
For calibrating NLI and duplicate detection models, we use input lengths (of premise/sentence1 and hypothesis/sentence2), softmax probabilities assigned to the labels, and the predicted label as the features.
We train calibrators using random forest implementations of Scikit-learn \cite{pedregosa2011scikit} for Calib C and Calib R approaches, and train a bert-base model for Calib T. 
In all calibration approaches, we calibrate using the IID held-out dataset and use softmax probability assigned to the positive class as the confidence estimate for SP.

\textbf{Label Smoothing: }For LS, we use MaxProb of the model trained with label smoothing as the confidence estimator for SP.
To the best of our knowledge, LS is designed for classification tasks only. Hence, we do not evaluate it for QA tasks.

\begin{figure}
    \centering
    \includegraphics[width=7cm]{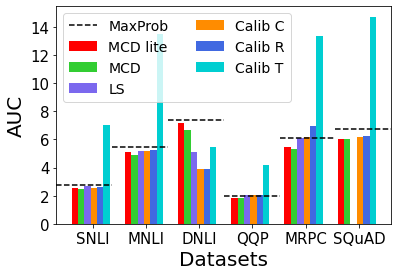}
    \caption{Comparing AUC of risk-coverage plot of various SP approaches with MaxProb in IID settings.}
    \label{fig:iid_results}
\end{figure}

\begin{figure*}[t]
\centering
    \begin{subfigure}{.1\textwidth}
        \includegraphics[width=\linewidth,height=25mm]{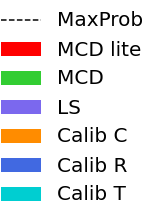}
    \end{subfigure}
    \begin{subfigure}{.37\textwidth}
        \includegraphics[width=\linewidth]{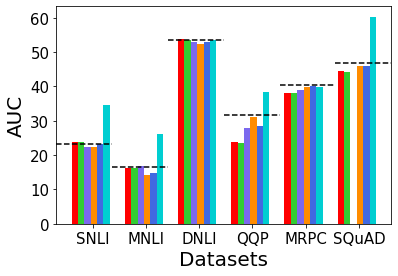}
        \caption{Out-Of-Domain}
        \label{ood_fig}
    \end{subfigure}
    \begin{subfigure}{.34\textwidth}
         \includegraphics[width=\linewidth,height=40mm]{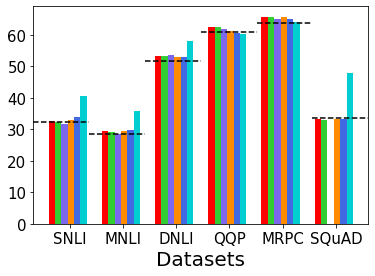}
         \caption{Adversarial}
         \label{adv_fig}
    \end{subfigure}
    
    \caption{Comparing AUC of risk-coverage plot of various approaches with MaxProb in OOD and ADV settings.
    The results have been averaged over all the task-specific OOD/ADV datasets mentioned in Section \ref{sec_experimental_setup} to highlight the general trend.
    Results of individual datasets have been provided in appendix.
    }
    \label{fig:ood_adv_results}    
\end{figure*}

\section{Results and Analysis}

\subsection{Slight Improvement in IID}
We compare SP performance of various approaches under IID setting in Figure \ref{fig:iid_results}.
Though all the approaches except Calib T outperform MaxProb in most cases, the magnitude of improvement is very small.
For instance, MCD achieves an average improvement of just $0.28$ on AUC value across all NLI datasets.

\textbf{Calib C and Calib R achieve the highest improvement on DNLI: }
\textit{We find that they benefit from using the predicted label as a feature for calibration.}
Specifically, the model's prediction accuracy varies greatly across labels ($0.94$, $0.91$, and $0.76$ for entailment, contradiction, and neutral labels respectively).
This implies that when the model's prediction is neutral, it is relatively less likely to be correct (at least in the IID setting). 
Calib C and R approaches leverage this signal and tune the confidence estimator using a held-out dataset and thus achieve superior SP performance.

\subsection{Negligible Improvement / Degradation in OOD and ADV}
Figure \ref{ood_fig}, \ref{adv_fig} compare the SP performance in OOD and ADV setting respectively.
The results have been averaged over all the task-specific OOD/ADV datasets mentioned in Section \ref{sec_experimental_setup} to observe the general trend.\footnote{Refer appendix for more details}
In the OOD setting, we find that the approaches lead to a negligible improvement in AUC. 
Notable improvement is achieved only by MCD in the case of QQP dataset.

\textbf{In ADV setting, all approaches degrade SP performance: }
\textit{Surprisingly, MCD that performed relatively well in IID and OOD settings, degrades more (by $1.74$ AUC) in comparison to other approaches} (except Calib T which does not perform well in all three settings).
This is because ensembling degrades the overall confidence estimate as the individual models of the ensemble achieve poor prediction accuracy in the ADV setting.

\subsection{Calib T Degrades Performance}
Calib C and Calib R slightly outperform MaxProb in most IID and OOD cases. 
However, Calib T considerably degrades the performance in nearly all the cases. 
\textit{We hypothesize that associating correctness directly with input text embeddings could be a harder challenge for the model as embeddings of correct and incorrect instances usually do not differ significantly. }
In contrast, as discussed before, providing features such as predicted label and softmax probabilities explicitly may help Calib C and R approaches in finding some distinguishing patterns that improve the selective prediction performance. 

\subsection{Existing Approaches Require Additional Resources}
Unlike typical ensembling, MCD does not require training or storing multiple models but, it requires making multiple inferences and can still become practically infeasible for large models such as BERT as their inference cost is high.
Calibration-based approaches need additional held-out data and careful feature engineering to train the calibrator.
\textit{Despite being computationally expensive, these approaches fail to consistently outperform MaxProb that does not require any such additional resources.}

\subsection{Effect of Increasing Dropout Masks in MCD}
With the increase in number of dropout masks used in MCD, the SP performance improves (from MCD lite with 10 masks $\rightarrow$ MCD with 30 masks).  
\textit{We hypothesize that combining more predictions on the same input results in a more accurate overall output due to the ensembling effect.}
However, we note that both MCD lite and MCD degrade SP performance in the ADV setting as previously explained.

\subsection{No Clear Winner}
None of the approaches consistently and considerably outperforms MaxProb in all three settings.
Most approaches do not fare well in OOD and ADV settings. 
Furthermore, a single approach does not achieve the highest performance across all tasks.
For instance, MCD outperforms all other approaches on Duplicate Detection datasets but does not perform well on NLI datasets (as Calib C beats MCD, especially in the OOD setting).
This indicates that \textit{these approaches do not translate well across tasks.}

\section{Conclusion}
Selective prediction ability is crucial for NLP systems to be reliably deployed in real-world applications and we presented the most systematic study of existing selective prediction methods.
Our study involves $17$ datasets across several NLP tasks and evaluation of existing selective prediction approaches in IID, OOD, and ADV settings.
We showed that despite leveraging additional resources (held-out data/computation), they fail to consistently and considerably outperform the simplest baseline (MaxProb) in all three settings.
Furthermore, we demonstrated that these approaches do not translate well across tasks as a single approach does not achieve the highest performance across all tasks.
Overall, our results highlight that there is a need to develop stronger selective prediction approaches that perform well across multiple tasks (QA, NLI, etc.) and settings (IID, OOD, and ADV) while being computationally efficient. 



\bibliography{anthology,custom}
\bibliographystyle{acl_natbib}
\newpage

\appendix

\section{Related Tasks}
\label{sec:related_tasks}
\subsection{Confidence Calibration}
\label{sec:background}
Selective Prediction is closely related to \textit{confidence calibration} \cite{platt1999probabilistic} i.e aligning model's output probability with the true probability of its predictions. 
Calibration focuses on adjusting the overall confidence level of a model, while selective prediction is based on relative confidence among the examples i.e systems are judged on their ability to rank correct predictions higher than incorrect predictions.

\subsection{Out-of-Domain Detection}
Using OOD Detection systems for selective prediction (abstain on all detected OOD instances) would be too conservative as it has been shown that models are able to correctly answer a significant fraction of OOD instances \cite{talmor-berant-2019-multiqa,hendrycks-etal-2020-pretrained, Mishra2020OurEM}.

\begin{figure*}[t]
\centering
    \begin{subfigure}{.4\textwidth}
         \includegraphics[width=\linewidth,height=50mm]{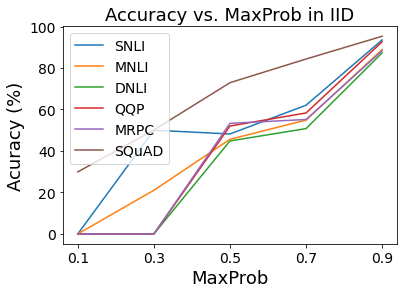}
         \caption{With increase in MaxProb, the accuracy usually increases.}
        \label{fig:maxProb_graph}
    \end{subfigure}
    \begin{subfigure}{.4\textwidth}
        \includegraphics[width=\linewidth,height=50mm]{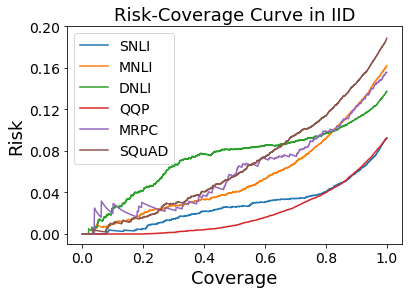}
        \caption{With increase in coverage (i.e decrease in abstention threshold), the risk usually increases.}
        \label{fig:rc_curves}
    \end{subfigure}
    \caption{Trend of Accuracy vs. MaxProb, Risk vs. Coverage for various models in the IID setting.}
    \label{fig:maxProb}    
\end{figure*}

\section{Why Lower AUC is Better?}
Small magnitude values of area under curve (AUC) are preferred as they represent low average risk across all confidence thresholds.

\section{Comparing SP Approaches}

Table \ref{tab:DD_SP_perf} compares SP performance (AUC of risk-coverage curve) of various approaches for Duplicate Detection datasets.
Table \ref{tab:QA_SP_perf} compares SP performance (AUC of risk-coverage curve) of various approaches for QA datasets.
Table \ref{tab:NLI_SP_perf} compares SP performance (AUC of risk-coverage curve) of various approaches for NLI datasets.
\begin{table}[t]
    \resizebox{\linewidth}{!}{
    \begin{tabular}{p{1.5cm}p{2cm}p{1cm}p{1.8cm}p{1.8cm}}
    \toprule
     \centering\textbf{Train On} &
        \textbf{Method}  & \textbf{IID$\downarrow$} &  \textbf{OOD avg.$\downarrow$}  & \textbf{ADV avg.$\downarrow$}\\
        \midrule
        
        \multirow{7}{*}{QQP} 
        & \textbf{MaxProb} & \underline{2.0} & \underline{31.72} & \underline{60.9} \\ 
         & MCD lite & 1.85 & \colorbox{green}{23.83} & 62.53 \\ 
         & MCD & \textbf{1.8} & \textbf{\colorbox{green}{23.61}} & 62.52 \\ 
         & LS & 2.08 & \colorbox{green}{27.92} & 61.92 \\ 
        & Calib C & 2.04 & 31.09 & 61.22 \\
        & Calib R & 2.07 & \colorbox{green}{28.53} & 60.68 \\ 
        & Calib T & 4.21 & 38.25 & \textbf{60.25} \\ 
         \midrule

        \multirow{7}{*}{MRPC} 
       & \textbf{MaxProb} & \underline{6.13} & \underline{40.46} & \textbf{\underline{63.88}} \\ 
     & MCD lite & 5.48 & \colorbox{green}{38.23} & 65.76 \\ 
     & MCD( & 5.35 & \textbf{\colorbox{green}{38.21}} & 65.62 \\ 
     & LS & 6.08 & 39.05 & 64.99 \\ 
     & Calib C & 6.17 & 39.82 & 64.99 \\ 
     & Calib R & 6.52 & 39.99 & 65.13 \\ 
     & Calib T & 13.35 & 39.75 & 64.22 \\

    \bottomrule
    \end{tabular}
    }
    \caption{
    Comparing selective prediction performance (AUC of risk-coverage curve) of various approaches for Duplicate Detection datasets. 
    Lower AUC is better in SP. 
    MaxProb baseline scores are \underline{underlined}, best performance is in \textbf{bold}, and scores that considerably outperform MaxProb are \colorbox{green}{highlighted}.
    }
    \label{tab:DD_SP_perf}
\end{table}
\begin{table}[t]
    \resizebox{\linewidth}{!}{
    \begin{tabular}{p{1.5cm}p{2cm}p{1cm}p{1.8cm}p{1.8cm}}
    \toprule
       \centering\textbf{Train On} & \textbf{Method}  & \textbf{IID$\downarrow$} &  \textbf{OOD avg.$\downarrow$}  & \textbf{ADV avg.$\downarrow$}\\
        \midrule
        
        \multirow{6}{*}{SQuAD} 
        & \textbf{MaxProb} & \underline{6.71} & \underline{46.73} & 33.69 \\ 
         & MCD lite & 6.06 & 44.56 & 33.34 \\ 
         & MCD  & \textbf{6.00} & \textbf{\colorbox{green}{44.35}} & 33.05 \\ 
        & Calib C & 6.15 & 45.93 & 33.27 \\ 
        & Calib R & 6.25 & 45.94 & 33.18 \\ 
        & Calib T & 14.72 & 60.31 & 47.87 \\

    \bottomrule
    \end{tabular}
    }
    \caption{
    Comparing selective prediction performance (AUC of risk-coverage curve) of various approaches for QA datasets. 
    Lower AUC is better in SP. 
    MaxProb baseline scores are \underline{underlined}, best performance is in \textbf{bold}, and scores that considerably outperform MaxProb are \colorbox{green}{highlighted}.
    }
    \label{tab:QA_SP_perf}
\end{table}
\begin{table}[t]
    \resizebox{\linewidth}{!}{
    \begin{tabular}{p{1.5cm}p{2cm}p{1cm}p{1.8cm}p{1.8cm}}
    \toprule
        \centering\textbf{Train On} & \textbf{Method}  & \textbf{IID$\downarrow$} &  \textbf{OOD avg.$\downarrow$}  & \textbf{ADV avg.$\downarrow$}\\
        
        \midrule
        \multirow{7}{*}{SNLI} 
        & \textbf{MaxProb} & \underline{2.78} & \underline{23.34} & \underline{32.4} \\ 
 & MCD(K=10) & 2.52 & 23.96 & 32.61 \\ 
 & MCD(K=30) & \textbf{2.47} & 23.81 & 32.47 \\ 
 & LS & 2.7 & \textbf{\colorbox{green}{22.42}} & \textbf{31.7} \\ 
 & Calib C & 2.57 & 22.47 & 33.0 \\ 
 & Calib R & 2.61 & 23.12 & 33.95 \\ 
 & Calib T & 7.02 & 34.74 & 40.68 \\ 
         \midrule

        \multirow{7}{*}{MNLI} 
      & \textbf{MaxProb} & \underline{5.47} & \underline{16.48} & \textbf{\underline{28.39}} \\ 
 & MCD(K=10) & 5.07 & 16.29 & 29.42 \\ 
 & MCD(K=30) & \textbf{4.92} & 16.18 & 29.18 \\ 
 & LS & 5.18 & 16.94 & 28.55 \\ 
 & Calib C & 5.16 & \textbf{\colorbox{green}{14.16}} & 29.57 \\ 
 & Calib R & 5.28 & {14.84} & 29.67 \\ 
 & Calib T & 13.51 & 26.12 & 35.79 \\

        \midrule
        
        \multirow{7}{*}{DNLI} 
        & \textbf{MaxProb} & \underline{7.36} & \underline{53.59} & \textbf{\underline{51.85}} \\ 
 & MCD(K=10) & 7.17 & 53.77 & 53.23 \\ 
 & MCD(K=30) & 6.69 & 53.67 & 53.24 \\ 
 & LS & \colorbox{green}{5.13} & 53.04 & 53.67 \\ 
 & Calib C & \textbf{\colorbox{green}{3.88}} & \textbf{52.35} & 52.91 \\ 
 & Calib R & \colorbox{green}{3.9} & 53.08 & 52.83 \\ 
 & Calib T & 5.46 & 53.58 & 58.13 \\     
  
    \bottomrule
    \end{tabular}
    }
    \caption{
    Comparing selective prediction performance (AUC of risk-coverage curve) of various approaches for NLI datasets. 
Lower AUC is better in SP. 
MaxProb baseline scores are \underline{underlined}, best performance is in \textbf{bold}, and scores that considerably outperform MaxProb are \colorbox{green}{highlighted}.
    }
    \label{tab:NLI_SP_perf}
\end{table}

\section{MaxProb for Selective Prediction}
Figure \ref{fig:maxProb_graph} shows the trend of accuracy against maxProb for various models in the IID setting.
It can be observed that with the increase in MaxProb the accuracy usually increases. 
This implies that a higher value of MaxProb corresponds to more likelihood of the model's prediction being correct.
Hence, MaxProb can be directly used as the confidence estimator for selective prediction.
We plot the risk-coverage curves using MaxProb as the SP technique in Figure \ref{fig:rc_curves}.
As expected, the risk increases with the increase in coverage for all the models.
We plot such curves for all techniques and compute area under them to compare their SP performance.
This shows that MaxProb is a simple yet strong baseline for selective prediction.

\section{Comparing Risk-Coverage Curves of MCD and Calib C for DNLI Dataset in IID Setting}
\begin{figure}
    \centering
    \includegraphics[width=7.6cm]{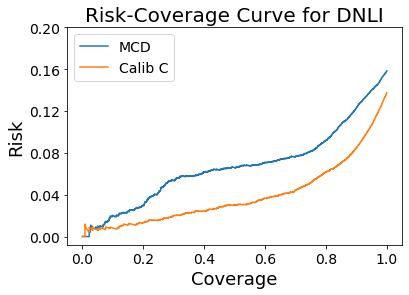}
    \caption{Comparing risk-coverage curves of MCD and Calib C for DNLI dataset in IID setting.}
    \label{fig:mcd_calib_analyze}
\end{figure}

We compare the risk-coverage curves of MCD and Calib C approaches on DNLI in Figure \ref{fig:mcd_calib_analyze}.
We observe that at all coverage points, Calib C achieves lower risk than MCD and hence is a better SP technique.
We find that they benefit from using the predicted label as a feature for calibration.
Specifically, the model's prediction accuracy varies greatly across labels ($0.94$, $0.91$, and $0.76$ for entailment, contradiction, and neutral labels respectively).
This implies that when the model's prediction is neutral, it is relatively less likely to be correct (at least in the IID setting). 
Calib C and R approaches leverage this signal and tune the confidence estimator using a held-out dataset and thus achieve superior SP performance.

\section{Composite SP Approach:}
We note that calibration techniques can be used in combination with Monte-Carlo dropout to further improve the SP performance. However, it would require even more additional resources i.e held-out datasets in addition to multiple inferences.



    
    
    

\end{document}